\documentclass{article}

    \PassOptionsToPackage{round}{natbib}

\usepackage{xcolor}
\usepackage[hidelinks]{hyperref}
\definecolor{linkcolor}{RGB}{0, 0, 128}
\hypersetup{
     colorlinks   = true,
     citecolor    = linkcolor,
     linkcolor    = linkcolor,
     urlcolor     = linkcolor,
}

     \usepackage[preprint]{neurips_2024}

\usepackage[utf8]{inputenc} 
\usepackage[T1]{fontenc}    
\usepackage{hyperref}       
\usepackage{url}            
\usepackage{booktabs}       
\usepackage{amsfonts}       
\usepackage{nicefrac}       
\usepackage{microtype}      
\usepackage{xcolor}         
\usepackage{multirow}
\usepackage{amsmath}
\usepackage{graphicx}
\usepackage{subcaption}
\usepackage{tcolorbox}
\usepackage{adjustbox}
\usepackage{todonotes}

\title{Improving Reward Models with Synthetic Critiques}

\author{%
  Zihuiwen Ye\thanks{This research was conducted during the author’s internship at Cohere.} \\
  University of Oxford\\
  \texttt{zihuiwen.ye@cs.ox.ac.uk} \\
  \And
  Fraser Greenlee-Scott \\
  Cohere \\
  \texttt{fraser@cohere.com} \\
  \And
  Max Bartolo \\
  Cohere \\
  \texttt{max@cohere.com} \\
  \And
  Phil Blunsom \\
  Cohere \\
  \texttt{phil@cohere.com} \\
  \And
  Jon Ander Campos \\
  Cohere \\
  \texttt{jonander@cohere.com} \\
  \And
  Matthias Gallé \\
  Cohere \\
  \texttt{matthias@cohere.com} \\
}

\begin{document}

\maketitle

\begin{abstract}

Reward models (RMs) play a critical role in aligning language models through the process of reinforcement learning from human feedback. RMs are trained to predict a score reflecting human preference, which requires significant time and cost for human annotation. Additionally, RMs tend to quickly overfit on superficial features in the training set, hindering their generalization performance on unseen distributions. We propose a novel approach using synthetic natural language critiques generated by large language models to provide additional feedback, evaluating aspects such as instruction following, correctness, and style. This offers richer signals and more robust features for RMs to assess and score on. We demonstrate that high-quality critiques improve the performance and data efficiency of RMs initialized from different pretrained models, reducing the reliance on costly human annotations. 
Furthermore, incorporating critiques improves both the interpretability and robustness of RM training. 
\end{abstract}

\section{Introduction}

Reinforcement learning from human feedback (RLHF) has emerged as a popular technique for aligning large language models (LLMs) with human preferences, underpinning the success of recent state-of-the-art LLMs \citep{ouyang2022training, bai2022training, achiam2023gpt}. RLHF typically involves two stages. First, a reward model (RM) is trained to generate scalar rewards based on human preferences. In the second stage, reinforcement learning algorithms
are employed to optimize language models by maximizing the rewards predicted by the trained RMs. 
The reward model plays a crucial role in the RLHF process, acting as a proxy for human assessment. 
In preference-based algorithms, it is used to estimate whether a user is likely to favor one text over another, built upon preference data collected from human annotators. Given an input instruction, the RM compares two completions and assigns a higher score to the preferred option.
Highlighting the significance of the reward model, Llama~2~\citep{touvron2023llama} utilized an extensive dataset of 1 million such binary preference datapoints to train their reward model, emphasizing its crucial role in achieving alignment with human preferences.

\begin{figure*}[ht]
 \centering \includegraphics[width=\linewidth]{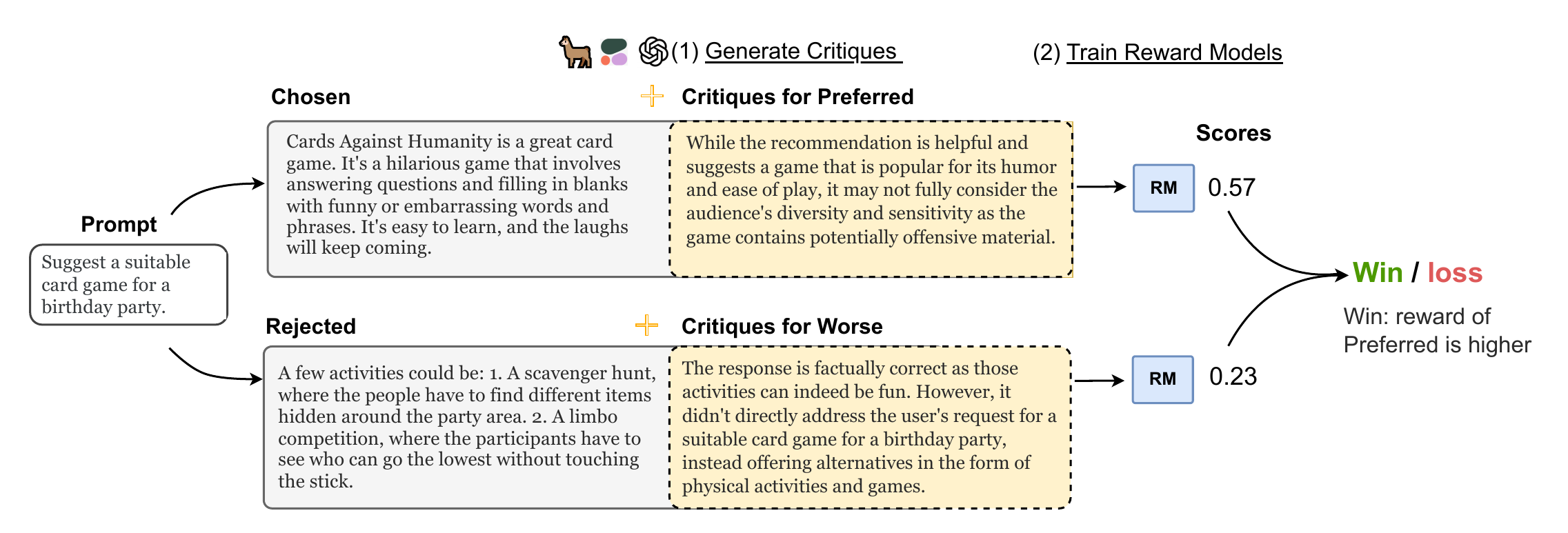}
 \vspace*{-5pt}
 \caption{A RM training example from \textsc{RewardBench} enriched with synthetic natural language critiques. We first prompt LLMs to generate critiques individually for each of the two prompt-completion pairs in a preference example, and then train a RM that predicts a scalar reward on top of them. The critiques break down both positive and negative features of the completion, evaluating it based on how effectively it fulfills the prompt requirements in aspects such as instruction-following, truthfulness, and helpfulness. \label{fig:diagram}}
\end{figure*}

Simultaneously, the prevailing approach to training preference-based reward models presents several challenges:
understanding how RLHF processes align with human preferences is challenging due to a lack of explainability and interpretability in RM training. 
Preference scores annotated by humans can be noisy, subjective and prone to biases \citep{hosking2023human, kirk2024prism}. 
Additionally, it is difficult to determine whether the RMs are fitting to superficial features or learning the actual preferences expressed by humans in the data, and consequently how such values are internalized by the LLMs during the second preference tuning stage \citep{casper2023open, marks2024training, qiu2024rethinking}. This process may also introduce biases, for example, favoring longer responses~\citep{singhal2023long}.

Reward models can also be brittle -- coupled with their tendency to pick up on data artefacts, this can lead to over-optimization towards the training distribution, resulting in inconsistencies between RM predictions and human preferences at inference time. Challenges also include generalization to distribution shifts, as well as susceptibility to adversarial attacks \citep{shen2023trickle, gao2023scaling, coste2023reward}. A final drawback is that training reward models requires training data, of a different kind than the one used to train generative text models.
Collecting these human preference annotations can be costly and laborious \citep{cui2023ultrafeedback}. 
Consequently, existing preferences datasets are often small in scale \citep{wu2023finegrained}, and limited on specific tasks \citep{stiennon2022learning}.

\smallskip

In this paper, we propose to use model-generated natural language critiques to overcome these limitations, where the critiques explicitly reason about the quality of a response. We illustrate our proposal in Fig.~\ref{fig:diagram}. In step (1), we prompt LLMs to generate synthetic critiques for each point-wise prompt-completion pair in the preference data. These critiques evaluate responses across multiple dimensions, including correctness and adherence to instructions, in a manner similar to Chain-of-Thought (CoT) reasoning  \citep{wei2022chain}. In step (2), we train RMs conditioned on these critiques to predict scalar rewards and evaluate their performance on a test set augmented with critiques. 
We hypothesize that critiques, by explicitly reasoning about the quality of responses and extracting textual features, can address the challenges posed by implicit and potentially ambiguous human annotations. We also aim to explore whether these critiques, by highlighting meaningful and task-relevant aspects, can enhance RM training efficiency and robustness.


We elaborate on our critique pipeline in the subsequent sections of the paper, with the objective of answering the following research questions:
\begin{enumerate}
\setlength\itemindent{0.5em} 
    \item[RQ1:] Do synthetic critiques improve RM performance? 
    \item[RQ2:] How do critiques impact RM performance and data efficiency as training scales up? 
\end{enumerate}

Our experiments show that using synthetic critiques improves RM performance, particularly in the low-resource data regime.
Notably, we show that a high-quality model-generated critique is roughly equivalent to 40 vanilla preference pairs, thus paving the way for more efficient use of annotation budgets. Additionally, we observe that critiques yield greater performance improvements on tasks requiring reasoning and adversarial skills, across various training data sizes. As synthetic critiques can be efficiently generated using open-source models, our method is both accessible and cost-effective.

\section{Related Work}

The current generation of LLMs \citep{touvron2023llama2,team2023gemini, achiam2023gpt, jiang2023mistral} is commonly trained using a multi-step process. This process usually consists of pretraining the model on a huge unsupervised corpus \citep{achiam2023gpt,le2023bloom,touvron2023llama}, supervised fine tuning on demonstrations \citep{zhang2023instruction} and reinforcement learning from human feedback \citep{christiano2017deep,ziegler2019fine,ouyang2022training}. The incorporation of human feedback has been essential for the success of such models \citep{ouyang2022training,chatgpt,achiam2023gpt,bai2022training}. 

\textbf{RLHF} usually involves training a reward model on preference data collected from annotators that is used for learning a policy using RL algorithms like \textsc{RLOO} \citep{ahmadian2024back} or \textsc{PPO} \citep{schulman2017proximal}. Different methods have been proposed for solving the reward modeling tasks. One of the most common ways to approach \textbf{reward modeling} is to apply the Bradley-Terry model \citep{bradley1952rank} for predicting human preference. \citet{rafailov2024direct} proposed Direct Preference Optimization, where instead of learning a separate reward model, the LLM can be used as a proxy reward model. 
Note that even these offline approaches require preference data, though without the need to train a separate reward model.
Lastly, there are many efforts that use generative LLMs as reward models by prompting the systems to choose a preference \citep{alpaca_eval,arenahard2024,zheng2023judging,verga2024replacing} or directly output a score \citep{cui2023ultrafeedback}. 

Due to the central role that reward models play in the RLHF process, there is an increasing interest for systematically understanding their behaviour. \citet{lambert2024rewardbench} introduced \textsc{RewardBench}, a benchmark dataset for evaluating reward models that contains pairs of completions with subtle but meaningful reasons of why one completion is preferred over the other. \textsc{RewardBench} demonstrated that while reward models are able to perform well on general chat, they struggle with more challenging reasoning examples. Apart from general performance, reward modeling is affected by issues such as length bias \citep{singhal2023long,shen2023loose} and lack of interpretability \citep{clymer2023generalization}. Moreover, these undesirable biases also mislead human annotators~\citep{hosking2023human}. 

\textbf{Critiques} have emerged as a method for improving interpretability during the model development process and robustness against adversarial examples \citep{wang2023pandalm}. 
These rationales have been proven to play an essential role during LLM training \citep{scheurer2023training, wu2023finegrained} and evaluation \citep{zheng2023judging, li2023generative}. Recent works explore using critiques for self-improvement by refining their outputs into better ones \citep{madaan2024self, gou2023critic, ye2023selfee}. Additionally, \citet{saunders2022selfcritiquing} demonstrate how critiques are helpful for human annotators in identifying flaws that they might have otherwise missed, while~\citet{yuan2024llmcrit} analyze the use of critiques to provide feedback to humans.

More similar to our work, \citet{zeng2023evaluating} find that when using language models as evaluators, prompting them with a rationale and rubric improves their evaluator accuracy. \textsc{Auto-J} \citep{li2023generative} extends this idea by training a generative LLM judge that is able to rate responses and generate critiques, improving the overall accuracy of the model. Concurrent works that explore using critiques for training RMs include \citet{ankner2024critique}. Despite critiques showing positive results in a wide variety of applications, their success heavily relies on their quality \citep{sun2024critique}. In order to measure critiques quality, \citet{lan2024criticbench} propose \textsc{CriticBench}, a benchmark designed to assess four dimensions of critique ability in LLMs: feedback, comparison, refinement, and meta-feedback.

\section{Methods}
In this section, we detail the method we use to generate synthetic critiques (\S\ref{sec:critique-generation}), and how we incorporate them during RM training (\S\ref{sec:critique-rm}).
\subsection{Synthetic Critique Generation with LLMs \label{sec:critique-generation}}
In the first step, we generate synthetic critiques for the preference data $\mathcal{D}$ used to train the RMs. 
We prompt LLMs to generate critiques as follows: Given a preference data pair $(x, y_+, y_-)$, where $x$ is a prompt and $(y_+, y_-)$ are the chosen and rejected completions, we prompt the LLM to generate point-wise critiques $c$ for each completion $y$. Specifically, for each chosen pair ($x$, $y$), where $y$ can be either a chosen or rejected completion, we generate a critique $c$. 
This way, we obtain a new critique-augmented training dataset with triplets $(x, y, c)$. We used the critique prompt from \citet{cui2023ultrafeedback}, designed to optimize feedback quality, to generate synthetic critiques $c$ that evaluate how well the completion meets prompt requirements, including instruction-following, truthfulness, and helpfulness. For details of the prompt template see App.~\ref{app:prompt_template}. An example of critique is shown in Fig.~\ref{fig:diagram}.

\subsection{Training RMs with critiques \label{sec:critique-rm}}
\paragraph{No-Critiques Baseline} We train reward models that take in a prompt and a completion to output a scalar score. We use a binary ranking loss following \citet{ouyang2022training}. Specifically, the loss function we use for No-Critiques baseline is:
\begin{equation}
\label{loss_rm_nocritique}
\mathcal{L(\theta}, \mathcal{D}) = - \mathbb{E}_{(x, y_{+}, y_{-}) \sim \mathcal{D}} 
 \log\left(\sigma\left(r_\theta(x,y_{+})-r_\theta(x,y_{-})-m(r)\right)\right)
\end{equation}
where $r_{\theta}(x, y)$ is the scalar output of the reward model for prompt $x$ and completion $y$ with trainable parameters $\theta$, and $m(r)$ is a margin that denotes the preference rating. A large margin is used for responses with distinct ratings, and a smaller one for similar responses. 
\paragraph{Critiques RM} Once we have the critiques $c$ generated from LLMs, we augment the training data by enriching them with critiques. We achieve this by concatenating the critiques after each completion to form new preference pairs ($x$, $y_{+}$ : $c_{+}$, $y_{-}$ : $c_{-}$)\footnote{In practice, we also add a short template describing the role of the following critique before $c$.}, which become a critique-augmented training set $\mathcal{D}'$.
We train critique RMs on $\mathcal{D'}$ with the loss:
\begin{equation}
\label{loss_rm_critique}
\mathcal{L(\theta}, \mathcal{D'}) = - \mathbb{E}_{(x, y_{+}, y_{-}, c) \sim \mathcal{D'}} 
 \log\left(\sigma\left(r_\theta(x,y_{+} : c_{+})-r_\theta(x,y_{-} : c_{-})-m(r)\right)\right)
\end{equation}

We implement our RMs by replacing the final projection layer of a pretrained language model with a linear layer that predicts a logit.
During inference, we similarly prompt for critiques using the same LLMs on the test set and use this critiques-augmented test set for evaluation. As a baseline, we compare with No-Critiques RMs trained on data without critiques added.

\section{Experimental Setup\label{sec:experimental_setup}}
In this section, we introduce the datasets we use (\S\ref{sec:datasets}), the LLMs for synthetic critiques generation (\S\ref{sec:critique_generator}), the pretrained models on which we build RMs (\S\ref{sec:rm_starting_ckpts}), and the training details (\S\ref{sec:implementation_details}).

\subsection{Datasets \label{sec:datasets}}

\begin{table}[t]
\small
\scalebox{0.85}{
\begin{tabular}{lll}
Dataset & \# Examples  &  Description  \\
\midrule
\textbf{RewardBench}   &   &   \\
\specialrule{0.01em}{1pt}{1pt}
Chat   & 358 & Open-ended chat prompts from AlpacaEval and MTBench. \\
Chat Hard   &  456 &  Questions from MTBench and LLMBar to stress test ability to understand trick questions. \\
Safety  &  740 & Prompts that test refusal of dangerous content from XSTest and Do-Not-Answer. \\
Reasoning &  1,431 & Code and reasoning prompts from HumanEvalPack and PRM800k.\\
\midrule
\textbf{PandaLM}   &  762   & Prompts from self-instruct, labeled by three independent human annotators.\\
\bottomrule
\end{tabular}
}
\vspace{0.5em}
\caption{Summary of the evaluation dataset used.}
\label{table:datasets}
\end{table}

For RM training, we collect a human-preference dataset that consists of 5k examples of open-ended, multi-turn conversations between a user and a chatbot. Each dataset entry includes a prompt or input instruction and two corresponding completions, with a human-annotated label indicating the better option. A preference rating is also given on a scale of three points (slightly better, better, significantly better). For more details of training data see App.~\ref{app:traindata_details}.
During inference, for each prompt paired with two completions from the test set, we use the trained RMs to assign a reward score for both the chosen and rejected completion. We use test accuracy as a metric to assess the performance of the RM, calculated based on the proportion of instances where the chosen completion receives a higher score than the rejected one.\footnote{We use accuracy on test prompts as a metric for evaluating RMs, under the premise that higher RM test accuracy correlates with enhanced downstream performance of LLMs trained with these RMs in RLHF.}

We evaluate the RMs on a variety of benchmark datasets that target a broad set of capabilities, including chat, instruction following, coding, and safety. A summary of the evaluation dataset is shown in Table~\ref{table:datasets}. Specifically, we evaluate on \textsc{Rewardbench} \citep{lambert2024rewardbench}, a dataset of prompt completion pairs that benchmarks how RMs perform on challenging, structured and out-of-distribution queries. \textsc{Rewardbench} contains four categories with a total of 2,985 prompts from different subsets: \textbf{Chat} (358 prompt in total) contains prompts from AlpacaEval \citep{alpaca_eval} and MTBench \citep{zheng2023judging} that test RM’s basic capability to distinguish correct responses in open-ended chat. \textbf{Chat Hard} (456) contains adversarial prompts from LLMBar \citep{zeng2023evaluating} and MTBench that stress test RM’s ability to understand trick questions and subtly different instruction responses, where the dispreferred output has appealing superficial qualities that challenge LLM-as-a-judge evaluators. \textbf{Safety} (740) contains pairs from XSTest \citep{rottger2023xstest} and Do-Not-Answer \citep{wang2023donotanswer} that test the model’s ability to refuse dangerous content and avoid incorrect refusals. \textbf{Reasoning} (1431) evaluates code and reasoning abilities. Code prompts from HumanEvalPack \citep{muennighoff2023octopack} have correct and buggy code as chosen  as rejected completions, and reasoning prompts are from PRM800k \citep{lightman2023let}. We also compute a \textbf{Avg. Score} by averaging the four categories with the same weighting as in \textsc{RewardBench}.

Additionally, we include PandaLM \citep{wang2023pandalm} as part of our evaluation set, consisting of 762 prompts. The instructions in this dataset are drawn from Self-Instruct~\citep{wang2022self}, with responses generated by different LLMs and each label independently provided by three different human evaluators.

\subsection{LLM Critique Generator \label{sec:critique_generator}}
To examine the effects of critiques, we select a range of LLMs with varying architectures, model sizes and training data for critique generation. We set up a pool of 6 models: (1) LLaMA2-7B-Chat \citep{touvron2023llama}, (2) LLaMA2-70B-Chat \citep{touvron2023llama}, (3) Mixtral-8x7B-Instruct \citep{jiang2024mixtral}, (4) Command R,\footnote{huggingface.co/CohereForAI/c4ai-command-r-v01} a 35B-parameter model, (5) Command R+\footnote{huggingface.co/CohereForAI/c4ai-command-r-plus} (103B), and (6) GPT4-Turbo\footnote{To be precise, we used \texttt{gpt-4-0125-preview}.}~\citep{achiam2023gpt}. 
Given a set of training and test preference data, we prompt each of the models to generate a set of critiques for both sets. For each experiment, we ensure that the train and test data match in that they are enriched by critiques generated by the same model. For baseline, we compare with No-Critiques, where we train and evaluate the RM on sets without critiques. 

\subsection{Pretrained Models for RM initialization \label{sec:rm_starting_ckpts}}
To explore the impact of pretrained model initialization for critiques RM training, we experiment with different pretrained model checkpoints to build the linear layer on top, varying in size and training data: (1) LLaMA2-7B-Base, (2) Command-35B-Base, and (3) Command R (35B). 
Command-35B-Base is a checkpoint from which Command R was obtained with further finetuning, and is what is generally called a ``base'' model.
Including Command R allows to check the impact that further supervised finetuning and preference modelling has on such a model.

\subsection{Training Details \label{sec:implementation_details}}
For all RMs in our experiments, we use a training batch size of 32 and train for 1 epoch with 155 steps in total, as we find that training for longer leads to overfitting. We train all the parameters, including both the pretrained model weights and the final linear layer. We use a cosine decay for learning rate schedule and Adam optimizer. The maximum learning rate is $8 \times 10^{-5}$ for all RMs starting on top of Command R and Command 35B-Base. The learning rate is decreased down to 10\% of the maximum, and we use a warmup of 32 steps. We run experiments on clusters of v4 TPUs. We observe that RM training starting from LLaMA2-7B-Base is sensitive to hyperparameters on training sets enriched with different model critiques. We hypothesize this sensitivity arises because LLaMA2-7B-Base is a relatively small base model that has not been instruction-tuned, making it susceptible to the varying distributions of critiques generated by different models. For this pretrained model, we conduct hyperparameter optimization on learning rate for RM runs enriched by different model critiques, on a separate validation set of 195 preference examples. We keep a cosine schedule with decay down to 10\%. We keep the warmup step as 5 and sweep the initial learning rate of $[5 \times 10^{-5}$, $8 \times 10^{-5}$, $1.6 \times 10^{-4}]$.

\section{Evaluation Results}

We train RMs on preference data enriched by synthetic critiques generated from different models starting from a range of pretrained model as detailed in \S\ref{sec:experimental_setup}. Here, we show the main findings of using critiques for RM training. Specifically, we answer RQ1 regarding the effectiveness of critiques in \S\ref{sec: critiques_accuracy}, RQ2 regarding the scaling behavior in \S\ref{sec:critiques_scaling}. We show finegrained result analysis in \S\ref{sec:discussion}.

\begin{table*}[ht]
\centering
\small
\scalebox{0.9}{
\begin{tabular}{l|cccc|c}
\toprule
\multicolumn{1}{c|}{\multirow{3}{*}{\textbf{Critiques}}} & \multicolumn{4}{c|}{\textbf{RewardBench}} & \multicolumn{1}{c}{\textbf{PandaLM}} \\
\cmidrule{2-6}
\multicolumn{1}{c|}{} 
& \textbf{Chat} & \textbf{Chat Hard} & \textbf{Safety} & \textbf{Reasoning}  & \textbf{Score}  \\
& (358) &  (456) &   (740) &  (1,431)  &  (762) \\
\midrule
\multicolumn{1}{l}{\textbf{LLaMA-7B Base Ckpt}} & \multicolumn{5}{c}{} \\
\midrule
No-Critiques &  0.2012 & 0.6425  &  0.6  &  0.5451  & 0.2507 \\
LLaMA2-7B-Chat  & \textbf{0.5674} & 0.4934 & 0.3878 & 0.431 & \underline{0.3976}  \\
Command R  & 0.2151 & 0.6557 & \underline{0.6041} & 0.5882   & 0.3123  \\ 
Mixtral-8x7B-Instruct  &  0.2147 & 0.6732 & 0.6 &  0.6678  & 0.3609  \\ 
LLaMA2-70B-Chat  & 0.206 & 0.6557 & 0.5878 & 0.6645   & 0.3898 \\
Command R+  &  0.1926 & \underline{0.6952}  & 0.6027 & \underline{0.6991}   & 0.3793  \\
GPT4-Turbo  & \underline{0.2452} & \textbf{0.6996} & \textbf{0.627}  & \textbf{0.7496}  & \textbf{0.4213} \\
\midrule
\multicolumn{1}{l}{\textbf{Command-35B Base Ckpt}} & \multicolumn{5}{c}{} \\
\midrule
No-Critiques &  0.2486 &  0.693  &  0.6  &  0.7178   & 0.3675  \\
LLaMA2-7B-Chat  & 0.2682 & 0.6952   &  0.5905  &  0.7394    & 0.4383 \\
Command R  &  0.2821 & 0.6667  &  \underline{0.6311}  &  0.7867   & 0.4383  \\
Mixtral-8x7B-Instruct  & 0.2989 & 0.6864 &  0.6014  &  0.7185  & 0.458 \\ 
LLaMA2-70B-Chat  &  0.2849 & 0.6732  & 0.6176  &  0.791  & 0.4501 \\
Command R+  & \underline{0.3073} & \underline{0.7149}  & 0.6176 & \underline{0.8144}   & \underline{0.5}  \\
GPT4-Turbo  & \textbf{0.3827} & \textbf{0.7961}  &  \textbf{0.6878}  &  \textbf{0.8527}   & \textbf{0.5066}\\
\midrule
\multicolumn{1}{l}{\textbf{Command R-35B Ckpt}} & \multicolumn{5}{c}{} \\
\midrule
No-Critiques & \textbf{0.3855} & 0.739  &  \textbf{0.7514}  &  \textbf{0.9033}   & \textbf{0.6339} \\
LLaMA2-7B-Chat  &  0.3352 & 0.7171 &  0.65 &  0.7453   & 0.5302 \\
Command R  & 0.3492 & 0.75 & 0.6527  &  \underline{0.879}   & 0.5433 \\
Mixtral-8x7B-Instruct  & 0.2709 &  0.7281  &  0.6351 &  0.8433  & 0.5144 \\ 
LLaMA2-70B-Chat  & 0.3603 &  0.7544  & 0.6324  &  0.8449  & \underline{0.5472} \\  
Command R+  & 0.3073 & \underline{0.7697}  &  0.6405  &  0.7277  & 0.5276 \\
GPT4-Turbo  & \underline{0.3799} & 0.\textbf{8246}  &  \underline{0.7405} &  0.8705   & 0.5131 \\
\midrule
\multicolumn{1}{l}{\textbf{Generative Judge Baseline}} & \multicolumn{5}{c}{} \\
\midrule
Gen-Command R+  & 0.5464 & 0.2765 & 0.6709 & 0.0673   & - \\
Gen-Command R+ (\textit{+Critiques}) & 0.5670 & 0.3635 & 0.7290 & 0.3685  & - \\
\midrule
\textbf{State-of-the-art}   &  &  &  &   \\
\midrule
TextEval-Llama3.1-70B & 0.941 & 0.901 & 0.932 & 0.964 \\
Llama-3.1-Nemotron-70B-Reward & 0.975 & 0.857 & 0.951 & 0.981 \\
\bottomrule
\end{tabular}
}
\vspace{1mm}
\setlength{\belowcaptionskip}{-0.5em}%
\caption{Test accuracy on \textsc{RewardBench} and PandaLM for RMs starting from three pretrained model checkpoints and enriched with various model critiques. Best results are bolded, and second-best are underlined. State-of-the-art results are taken from the \textsc{RewardBench} leaderboard (Oct 2024).} 
\label{tab:test_accuracy}
\end{table*}

\subsection{Do synthetic critiques improve RM performance? \label{sec: critiques_accuracy}}

\begin{figure}[t]
\centering \includegraphics[width=0.65\linewidth]{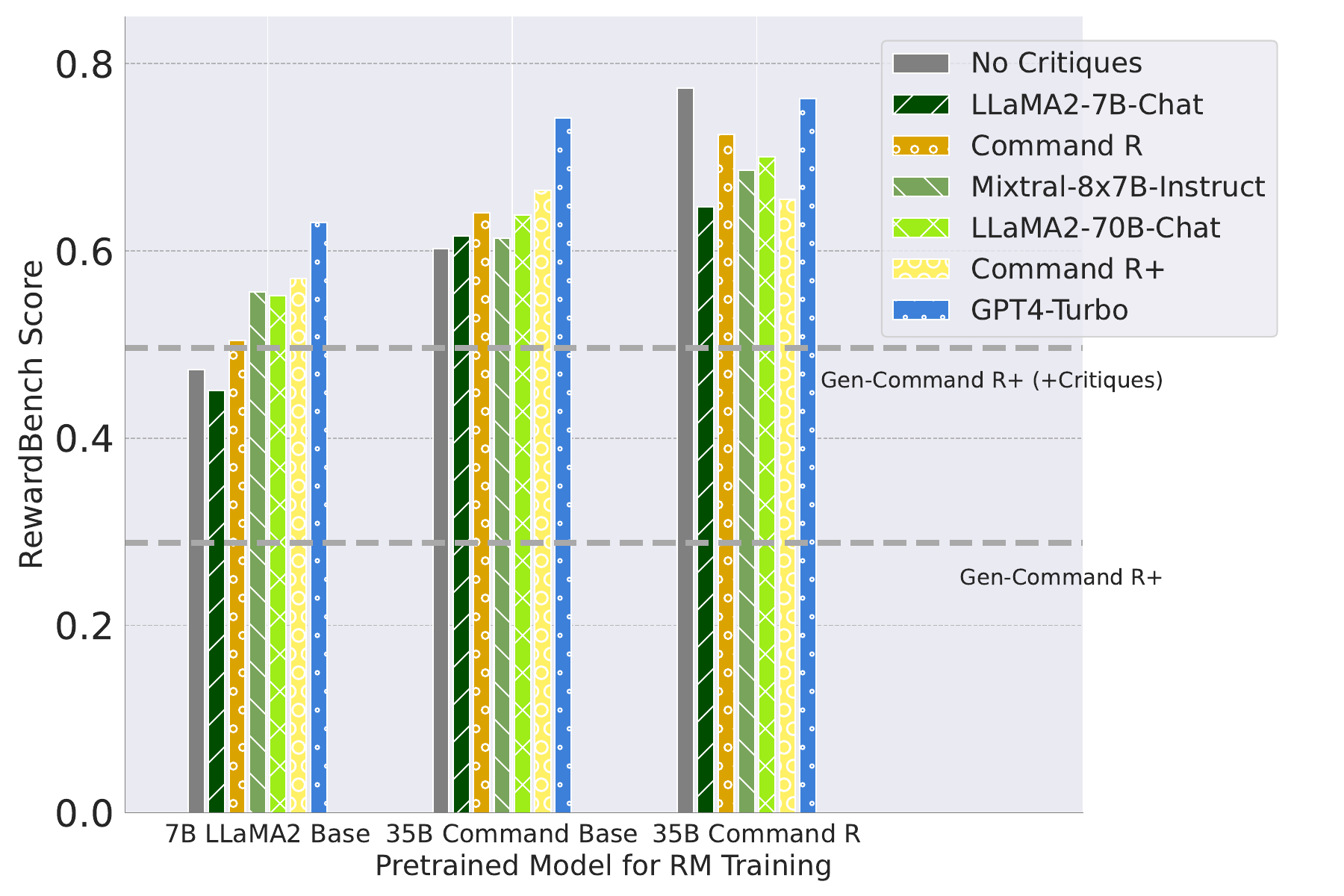}
 \vspace*{-5pt}
 \caption{\textbf{Avg. Score} on \textsc{RewardBench}, which shows the weighted average test accuracy of each category under \textsc{RewardBench}. Strong critiques, such as those generated by GPT4-Turbo, improve scores on RMs starting from 7B LLaMA2 Base and 35B Command Base, whereas weaker critiques, such as those generated by LLaMA2-7B-Chat, could be detrimental. 35B Command R is a strong preference tuned model for RM on which No-Critiques excel.\label{fig:rewardbench_score}}
\end{figure}

Table \ref{tab:test_accuracy} summarizes the evaluation results on the four subsets in \textsc{RewardBench}, and PandaLM. Fig.~\ref{fig:rewardbench_score} displays \textsc{RewardBench} \textbf{Avg. Score}. In most cases, adding critiques improves RM test accuracy compared to No-Critiques baseline, illustrating the effectiveness of synthetic critiques. 

\paragraph{High-quality critiques improve RM test accuracy.} We observe that critique quality has an impact on RM test performances. Specifically, across the base pretrained models, stronger models such as GPT4-Turbo consistently provide the highest \textbf{Avg. Score} for both datasets, with a score increase of 0.16 and 0.14 on \textsc{RewardBench} for RMs starting from LLaMA-7B Base and Command-35B Base respectively. This is followed by Command R+, with an increase of 0.1 and 0.062 respectively. On the other hand, weaker critiques generated from a smaller model such as LLaMa2-7B-Chat have a detrimental effect on test accuracy on LLaMA-7B Base. 
Similar to \citet{saunders2022selfcritiquing} and \citet{sun2024critique}, we observe a positive correlation between the effect of critiques on test performance and critique model sizes, as shown in Fig.~\ref{fig:rewardbench_vs_sizes}. For more quantitative and qualitative analysis of critique quality with metacritique scores~\citep{sun2024critique}, see App.~\ref{app:critique_quality}. 

\paragraph{Critiques are more effective on weaker checkpoints. \label{sec:starting_ckpts}} We found that the choice of the pretrained model for RM also plays a role in the final performance. Comparing across pretrained checkpoints, we observe that RM performance improves with the capacity of the starting models, both without critiques and with critiques. Notably, critiques significantly enhance performance in RMs starting from weaker base models, such as LLaMA2-7B Base. On the other hand, the Command R-35B model performs well across tasks even without any critiques, notably in \textit{Safety} and \textit{Reasoning}. We speculate that this is because Command R-35B, unlike the other base models, has already been trained on a large amount of preference data, giving it strong preference modeling capabilities before RM training starts, whereas starting from base models provides more room for RMs to leverage synthetic critiques effectively in predicting their final reward.

\subsection{What impacts do critiques have on RMs as training scales up?\label{sec:critiques_scaling}}
Next, we conduct experiments to study the scaling behavior of RM training with critiques. We prepare four No-Critiques training datasets with 5k, 50k, 100k, and 200k preference examples. We then generate synthetic critiques for all four sets and train RMs on them to compare with No-Critiques.

\begin{figure}[t]
 \centering \includegraphics[width=0.78\linewidth]{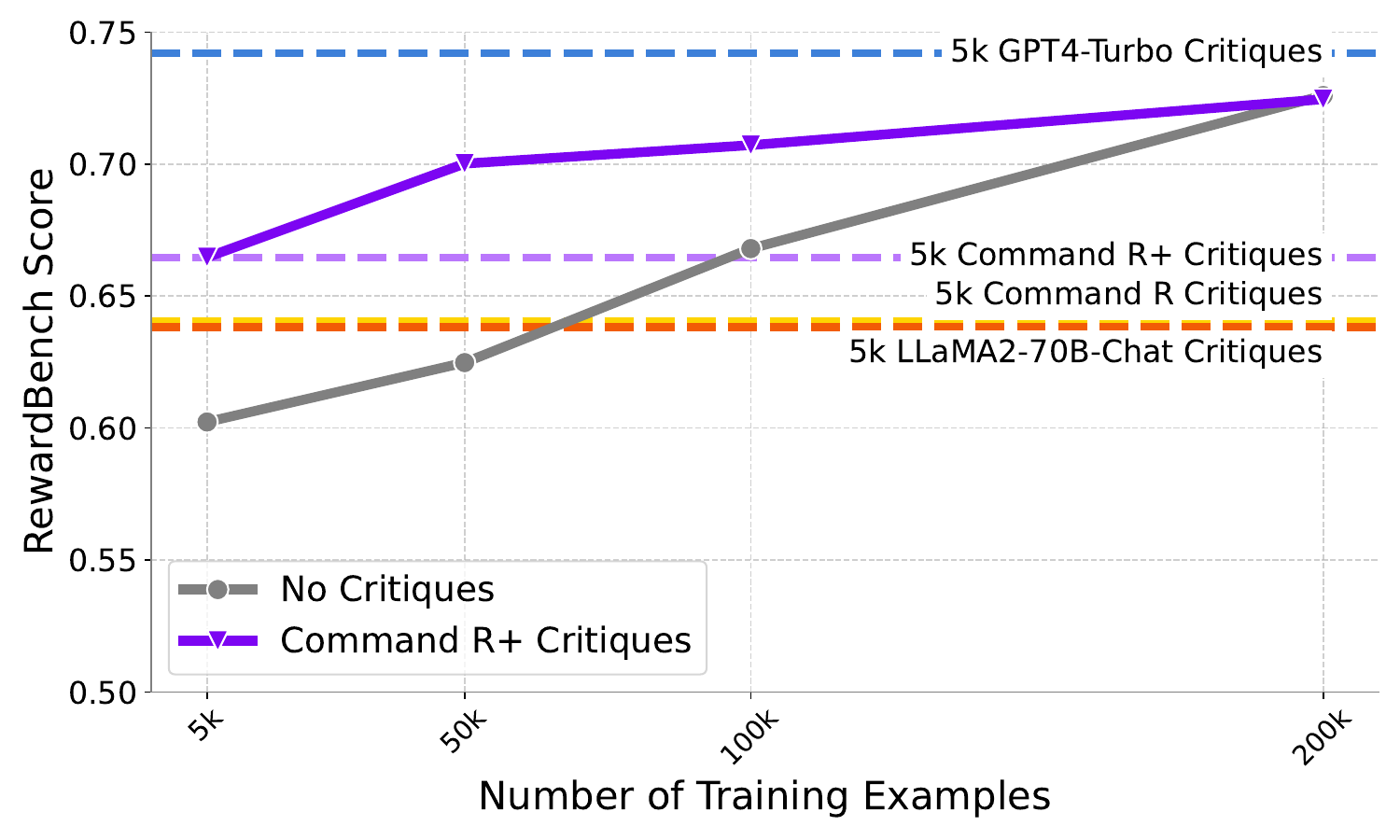}
 \vspace*{-5pt}
 \caption{\textsc{RewardBench} Score scaling behavior for RMs trained with increasing number of training examples, with No-Critiques and Command R+ Critiques. Using critiques enhances data efficiency, particularly in low-data setting. Dashed lines show scores achieved by 5k synthetic critiques, which reach a comparable No-Critiques test accuracy with significantly less data. \label{fig:scaling_35b_base}}
\end{figure}

\paragraph{Critiques Increase Data Efficiency.} We initialize RMs from the 35B Command R base model and generate synthetic critiques using Command R+. We use \textsc{RewardBench} \textbf{Avg. Score} for evaluation, as illustrated in  
Fig.~\ref{fig:scaling_35b_base}. We observe that on \textsc{RewardBench}, increasing the number of training examples generally increases test accuracy, both with and without critiques. Using synthetic critiques enhances data efficiency, as demonstrated by the greater accuracy gains of Command R+ Critiques over No-Critiques, particularly in low-data settings (5k-50k). The benefit of critiques becomes less pronounced with abundant data, as both RMs achieve comparable test performance around 200k training examples. However, when critique quality is high, we continue to observe a substantial performance boost even in abundant-data settings: the performance gap remains evident at 200k examples, with just 5k strong critique models (dashed line of GPT-4 in Fig.~\ref{fig:scaling_35b_base}).

\begin{figure}[t]
    \small
    \centering
    \begin{subfigure}[b]{0.48\textwidth}
        \centering
        \includegraphics[width=\textwidth]{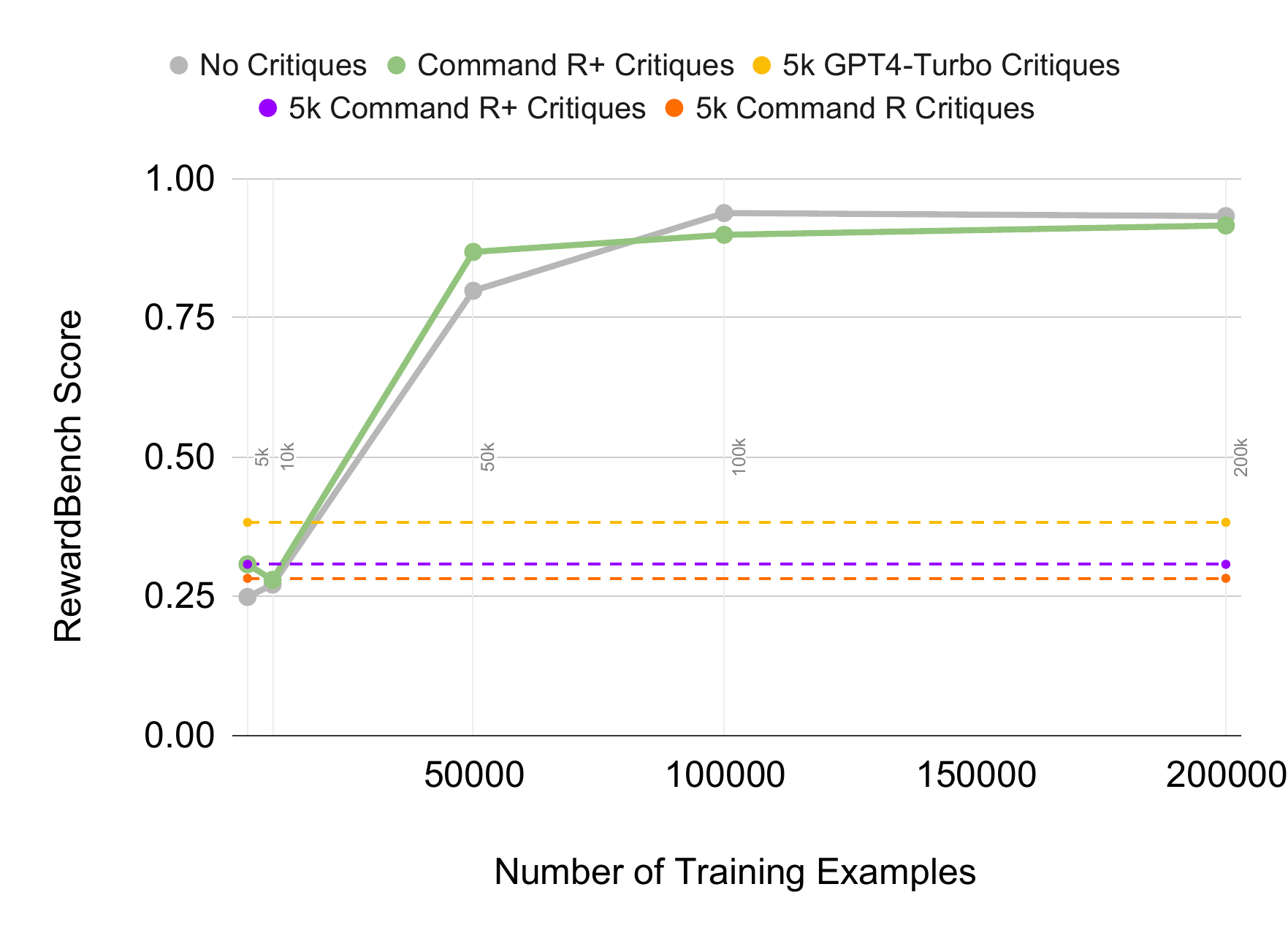}
        \caption{Chat}
        \label{fig:rbsubsets_chat}
    \end{subfigure}
    \hfill
    \begin{subfigure}[b]{0.48\textwidth}
        \centering
        \includegraphics[width=\textwidth]{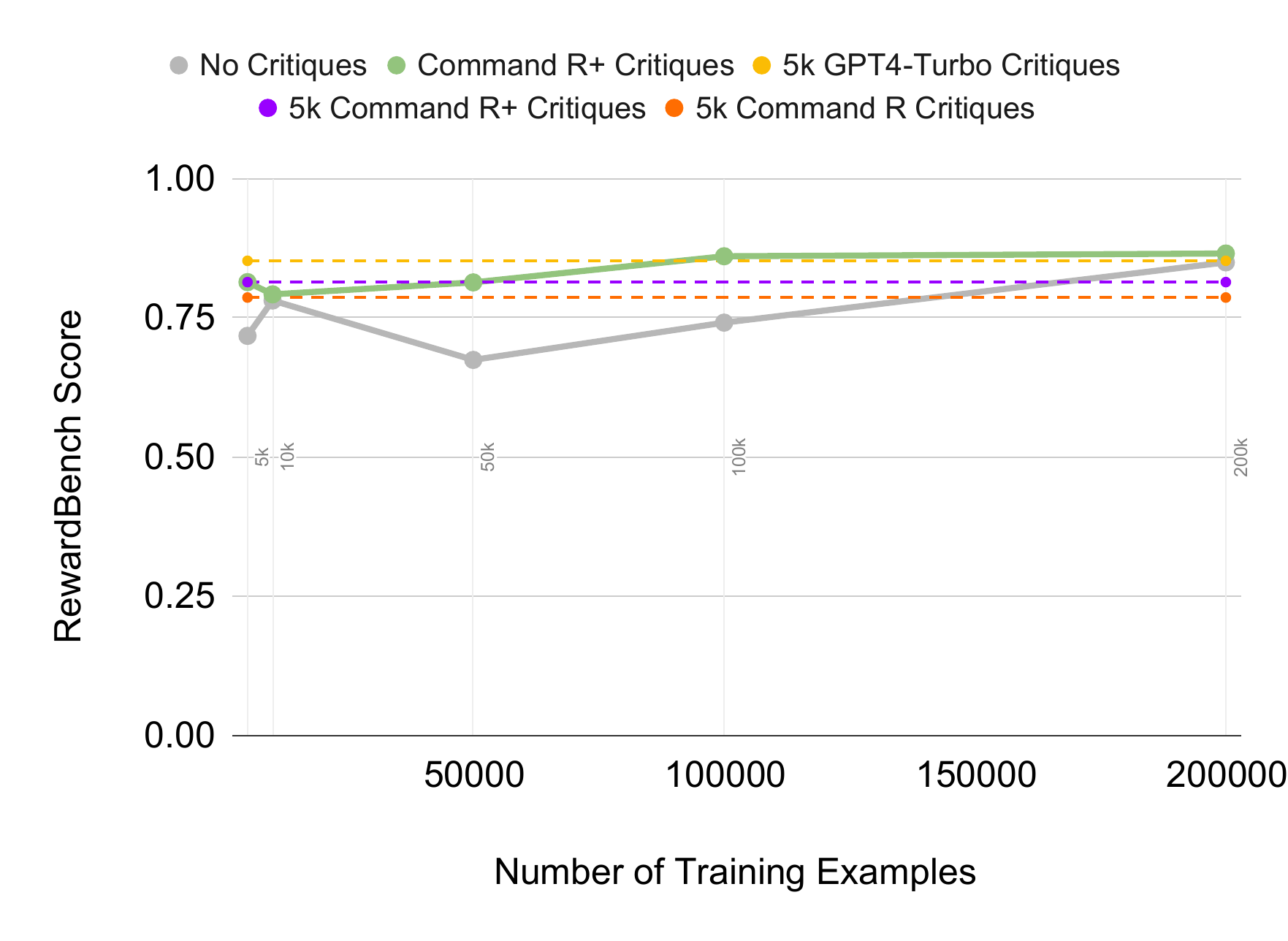}
        \caption{Reasoning}
        \label{fig:rbsubsets_reasoning}
    \end{subfigure}
    \caption{Accuraccy for \textit{Chat} and \textit{Reasoning} section of \textsc{RewardBench} for RMs trained with increasing number of training examples, with No-Critiques and Command R+ Critiques.}
    \label{fig:rbsubsets}
\end{figure}

\paragraph{How many No-Critiques examples Is a Critique Worth?} To illustrate the effectiveness of strong critiques, we plot the \textsc{RewardBench} score from Table \ref{tab:test_accuracy} for RMs trained with 5k synthetic critiques using Command R, LLaMA2-70B-Chat, Command R+, and GPT4-Turbo as dashed lines in the background of Fig.~\ref{fig:scaling_35b_base}. We observe that 5k Command R+ Critiques achieve a test accuracy comparable to approximately 90k No-Critiques. Similarly, 5k Command R and 5k LLaMA2-70B-Chat equate to roughly 62k No-Critiques, while 5k GPT4-Turbo surpass even 200k No-Critiques. 

\subsection{Analysis}
\label{sec:discussion}

\paragraph{Finegrained Analysis}
We observe that the benefit of critiques varies by task type. We present a detailed breakdown of test accuracy over different tasks under \textsc{RewardBench} with scaling in Fig.~\ref{fig:rbsubsets_reasoning} and App.~~\ref{app:performance_by_task}. As shown in Table~\ref{tab:test_accuracy}, the accuracy on the \textit{Chat} subset of \textsc{RewardBench} is low (often below random chance) with 5k examples. However, Fig.~\ref{fig:rbsubsets_chat} demonstrates a significant improvement when scaling to 10k pairs, achieving very high scores (0.91) and potentially saturating this task.
Compare this to Fig.~\ref{fig:rbsubsets_reasoning} on \textit{Reasoning} -- a harder task -- where using critiques outperforms the baseline (by more than 10 points) and where they only converge with 20k training examples. In addition, Fig.~\ref{fig:rbsubsets_chathard} shows that training with critiques leads to greater robustness against adversarial examples on \textit{Chat Hard}, and the effect is persistent across training data sizes. Interestingly, critiques offer a more significant performance boost for tasks involving reasoning and adversarial skills, across increasing training data. We hypothesize that this is attributed to the Chain-of-Thought style feedback within critiques, which provides additional context for RMs to condition on when predicting a scalar score. We leave it as future work to explore the impact of critiques on LLM reasoning abilities when preference-tuned with such critique-enhanced RMs.

\paragraph{Critiques versus Knowledge Distillation}
As the critiques in our experiments are generated from another model, training critiques RM can be seen as a form of knowledge distillation. To understand the benefits of critiques versus general knowledge transfer, we perform baseline studies where we directly prompt the critique generator for Likert scores (on a scale of 0 to 10), 
for each point-wise prompt-completion pair in the test set, and compute the proportion that is correct if the generated score is higher on the preferred completion. This setting resembles the performance of an inference-time generative RM \citep{li2023generative} and can serve as an upper bound of the test set performance achievable from a distilled model. We call this baseline Generative Judge RM and present the results with a Command R+ critique generator (called Gen-Command R+). We further include another inference-time baseline, Gen-Command R+ (\textit{+Critiques}) where we additionally prompt for a critique before outputting a score. We use the same prompt template for critiques RMs in App.~\ref{app:prompt_template}. In Table \ref{tab:test_accuracy}, we observe that while Gen-Command R+ performs strongly on \textit{Chat} and comparably on \textit{Safety}, the critiques RMs, across different critique generators, overwhelmingly outperform this baseline on \textit{Reasoning} and \textit{Chat Hard}, validating their effectiveness particularly in more challenging and adversarial prompts. For Gen-Command R+ (\textit{+Critiques}), we observe that similar to training scalar RMs, conditioning on critiques during inference leads to a higher test score (with an increase of 0.3 on \textit{Reasoning}) than No-Critiques, agreeing with the observations in \citet{wang2023pandalm}. As the benefit of critiques exists in both training and inference, using critiques is not merely a form of knowledge transfer but introduces an additional Chain-of-Thought that enhances the model's decision-making process.

\section{Conclusion}
\label{sec:conclusion}
We propose an accessible and cost-effective approach to improve reward models (RMs) in RLHF by enriching them with synthetic natural language critiques. 
We prompt LLMs to generate critiques that assess the prompt-completion pair and train RMs to predict scalar rewards conditioned on these critiques. 
Experiments demonstrate the effectiveness of using critiques for RMs, as evidenced by their stronger performance on existing RM benchmarks. 
We found that critiques are especially effective in low-data settings, when they are of high quality, the base RM is weak, and the task requires reasoning skills. In these cases, a single high-quality critique-enhanced preference pair can be as valuable as 40 standard preference pairs.
As these critiques are generated without human labour, this approach offers a more cost-effective way to obtain competitive reward models, given the current scarcity and high cost of preference training data. We also show that our method of leveraging critiques as part of the reward model’s Chain-of-Thought enhances the training process and interpretability. In future work, we plan to investigate their potential to improve LLM reasoning abilities.

\section{Limitations}
While we evaluate RMs using ranking test accuracy, we did not experiment with optimizing LLMs with such critique-enriched RMs, but focused on evaluating RMs directly on existing RM benchmarks. However, previous work \citep{shen2023trickle} suggests that stronger RMs lead to enhanced performance of LLMs during RLHF. Regarding critique quality, the critiques generated by the LLMs might be incorrect, or contain hallucinations. 
As mentioned in \S~\ref{sec:critiques_scaling}, the added value of critiques diminishes in the abundant data regime. 
However, even in that scenario adding critiques does not reduce the final performance and obtains better score in the more challenging settings such as \textit{Reasoning} and \textit{Chat Hard}, indicating better generalization.
A final limitation is the limited improvement when starting from a strong preference tuned model, which is similar to the case of abundant preference data.
We would also like to extend our experiments to include a wider set of pretrained checkpoints, and prompting methods to understand their effects on critiques.

\begin{ack}
We extend our gratitude to those who provided invaluable feedback and suggestions for this paper. We would like to especially thank Sander Land, Maximilian Mozes, Nathan Grinsztajn, Irem Ergün, and Andy Toulis for helpful discussions and assistance with the implementation. We would also like to thank our colleagues at Cohere for their continued support throughout this project.
\end{ack}

\bibliographystyle{plainnat}
\bibliography{references}

\appendix


\section{Prompt Template \label{app:prompt_template}}

We show the template we use to prompt LLMs for point-wise critiques in Fig.~\ref{fig:critique_template}. We made minimal changes starting from the template in \citet{cui2023ultrafeedback}.
\begin{figure*}[!th]
\begin{tcolorbox}[
    colback=white, 
    colframe=black, 
    title=\textbf{Template for Point-wise Critique Generation}, 
    fonttitle=\bfseries\large, 
    arc=4mm, 
]
User: \{prompt\}

Chatbot: \{response\} \textit{\# could be either chosen or rejected}

Please provide a critique of the last response in one short paragraph. Your critique should be concise, specific, insightful and to the point. Aspects you should consider are: (1) Helpfulness. A good response should precisely/closely answer the user’s request. (2) Correctness. A good response should be honest and factually correct.
\end{tcolorbox}
\end{figure*}\label{fig:critique_template}

\begin{table}[t]
\small
\begin{tabular}{@{}p{\textwidth}@{}}
\toprule
\textbf{User:} Brainstorm uses for a paperclip. \\ 
\midrule
\begin{tabular}{@{}p{\textwidth}@{}}
\textbf{preferred response}: A paperclip can be used to keep papers together, hang items on a bulletin board, mark pages in a book, or create a makeshift zipper pull. It can also be used to repair torn clothing or to attach items to a keychain.
\end{tabular}\\
\midrule
\begin{tabular}{@{}p{\textwidth}@{}}
\textbf{LLaMA2-7B-Chat}: The response provided is helpful in addressing the user's request by providing a list of potential uses for a paperclip. However, the list is quite general and does not provide any particularly creative or innovative ideas for using a paperclip. Additionally, some of the listed uses, such as repairing torn clothing, are not necessarily unique to paperclips and could be accomplished with other materials as well. Overall, the response is factually correct but could benefit from more specific and creative suggestions for using a paperclip. \\
\midrule
\textbf{Command R}: The suggestions provided for alternative uses of a paperclip are concise and closely aligned with the user's request. However, some of the uses, such as using a paperclip for keeping papers together or fastening items to a keychain, are relatively obvious and common. While the response is factually correct, it may not offer innovative or unexpected ideas, falling short of providing a truly insightful or surprising critique. \\
\midrule
\textbf{Mixtral-8x7B-Instruct}: The response is helpful in providing a list of potential uses for a paperclip, but it could be improved by providing more detail or context for each use case. For example, instead of simply stating that a paperclip can be used to mark pages in a book, the response could explain how to do this or provide tips for using a paperclip in this way. Additionally, the response could be more specific in terms of the types of items that can be attached to a keychain using a paperclip. Overall, while the response is factually correct, it could be more helpful by providing more detailed and specific information. \\
\midrule
\textbf{LLaMA2-70B-Chat}: The previous response provides some useful ideas for using a paperclip, such as keeping papers together, marking pages in a book, and creating a makeshift zipper pull. However, it falls short in terms of creativity and imagination. The suggestions offered are fairly obvious and do not fully explore the potential uses of a paperclip. Additionally, the response does not provide any specific examples or explanations to help the user better understand how to implement these ideas. Overall, while the response is not incorrect, it could be more helpful and inspiring for the user. \\
\midrule
\textbf{Command R+}: The response provides a concise and varied list of possible paperclip uses, offering an insightful and helpful critique of the everyday item's potential. The suggestions are creative and practical, covering a range of scenarios where a paperclip could be utilized. The ideas are correct and sensible, and the tone is positive and straightforward, closely answering the user's request.\\
\midrule
\textbf{GPT4-Turbo}:  The last response, while providing a brief overview of potential uses for a paperclip, falls short in terms of depth and creativity. It lists a few common uses but misses the opportunity to explore more innovative or less conventional applications, such as resetting electronic devices through their pinhole reset buttons, using it as a SIM card eject tool in smartphones, or even bending it into a stand for small items like business cards or phones. The response is helpful and correct in what it does offer, ensuring factual accuracy in the suggested uses. However, it could significantly benefit from expanding the scope of applications to fully address the user's request for brainstorming uses, thereby enhancing its helpfulness and engaging the user's curiosity more effectively.\\ 
\end{tabular}
\\
\bottomrule
\end{tabular}
\caption{Example of point-wise critiques generated by different models for a preferred preference response.}
\label{tab:case_ppo}
\end{table}

\begin{figure}[h!]
\centering
\fbox{
\begin{minipage}{0.95\linewidth}
\textbf{Prompt}: Write a poem about the ocean.
\vspace{1em}

\textbf{Chosen}: Sure! Here is a poem about the ocean: The ocean, a majestic mystery, A universe of its own, yet succinctly, A force that brings life, calm and strife, Ever-changing, a constant revolution. Powerful, yet gentle, peacefully still, A mirror, a dancer, a rousing thrill. Blue like the sky, a vivid, soulful hue, A home to creatures, big and small, novel. From mighty whales to tiny, shimmering fish, The ocean's depths unfold with a rush, A treasure trove of diversity, Its secrets yet to be unlocked, unfurled. With waves that caress, then attack with might, The ocean's embrace is an exhilarating light, A force that shapes our fragile Earth, A magical, ethereal, breathtaking worth. But like a tempest, it can roar and rage, Sending fear into the heart of the age, Unleashing chaos, a devastating spell, A reminder of nature's strength, supreme. The ocean, a paradox, serene and fierce, A vibrant, vital force, a beautiful deity. Its call echoes through time, a yearning heard, A vast, ethereal wonder, a mystery unfurled. And in its depths, the soul finds a home, A place of wonder, a place of profound solace and roam. Does this poem capture the essence of the ocean for you? Let me know if you would like me to make any adjustments or incorporate any other specific themes or ideas about the ocean into the poem.

\vspace{1em}

\textbf{Rejected}: The ocean's mystery unfolds, A story that begins where land meets shore. Beneath its surface, a new world is found, A captivating haven, a haven to be explored. Its vastness stretches far and wide, A home to countless creatures, big and small, From mighty whales to tiny plankton, Each playing a part in this oceanic pantomime. The ocean's rhythm is ancient and true, A flow that ebbs and flows with every moon, It whispers secrets of a bygone era, A symphony of reminders of love that it did confer. Its power is mighty and commanding, Yet it can be as gentle as a caress, It nourishes and sustains, A source of life, a mirror of our deepest wishes. It is a mirror of our souls, A reflection of our inner selves, A place of discovery, Where the secrets of the universe are finally revealed. The ocean's mystery is a story never told, But written with each wave's constant toil, It is a part of us, And a part of the world that is ever-evolving. And though its secrets may seem reserved, They are there for those who are willing to explore, To those who yearn to discover more, The ocean's mystery is a tale to be uncovered, not ignored.

\vspace{1em}
\textbf{Confidence}: Slightly Better

\end{minipage}
}
\caption{An example from the training dataset.}
\label{fig:train_example}
\end{figure}

\section{Training Data Format \label{app:traindata_details}}
We detail the data collection process for our training preference data, which contain multi-turn conversational tasks seeded with an initial prompt that is synthetically generated. 
Two completions are sampled from a previous generation of a model, and the preference label is provided by one human annotator. The human annotators were paid more than minimal wage for the annotation task. This dataset does not contain any tie, so there is always one completion which is considered better than the other.
To define how much it is each pair is also annotated with a confidence margin (slightly better, better, significantly better), which indicates how confident the annotator is about the ranking decision. 
The distribution is as followed in increasing confidence margin: 86.6\% slightly better, 12.6\% better, 0.8\% significantly better. In Figure~\ref{fig:train_example} we show an example prompt and completion pairs, along with the label and confidence margin.
For the evaluation dataset, the \textsc{RewardBench} \citep{lambert2024rewardbench} dataset and PandaLM \citep{wang2023pandalm} contain subsets of various license types and can be found respectively at: \url{https://huggingface.co/datasets/allenai/reward-bench}, \url{ https://github.com/lmmlzn/Awesome-LLMs-Datasets}.

\begin{figure}[t]
    \centering
    \begin{subfigure}[b]{0.49\textwidth}
        \centering
        \includegraphics[width=\textwidth]{figures/reasoning.pdf}
        \caption{Reasoning}
        \label{fig:subplot1}
    \end{subfigure}
    \hfill
    \raisebox{0.0cm}{
    \begin{subfigure}[b]{0.49\textwidth}
        \centering
        \includegraphics[width=\textwidth]{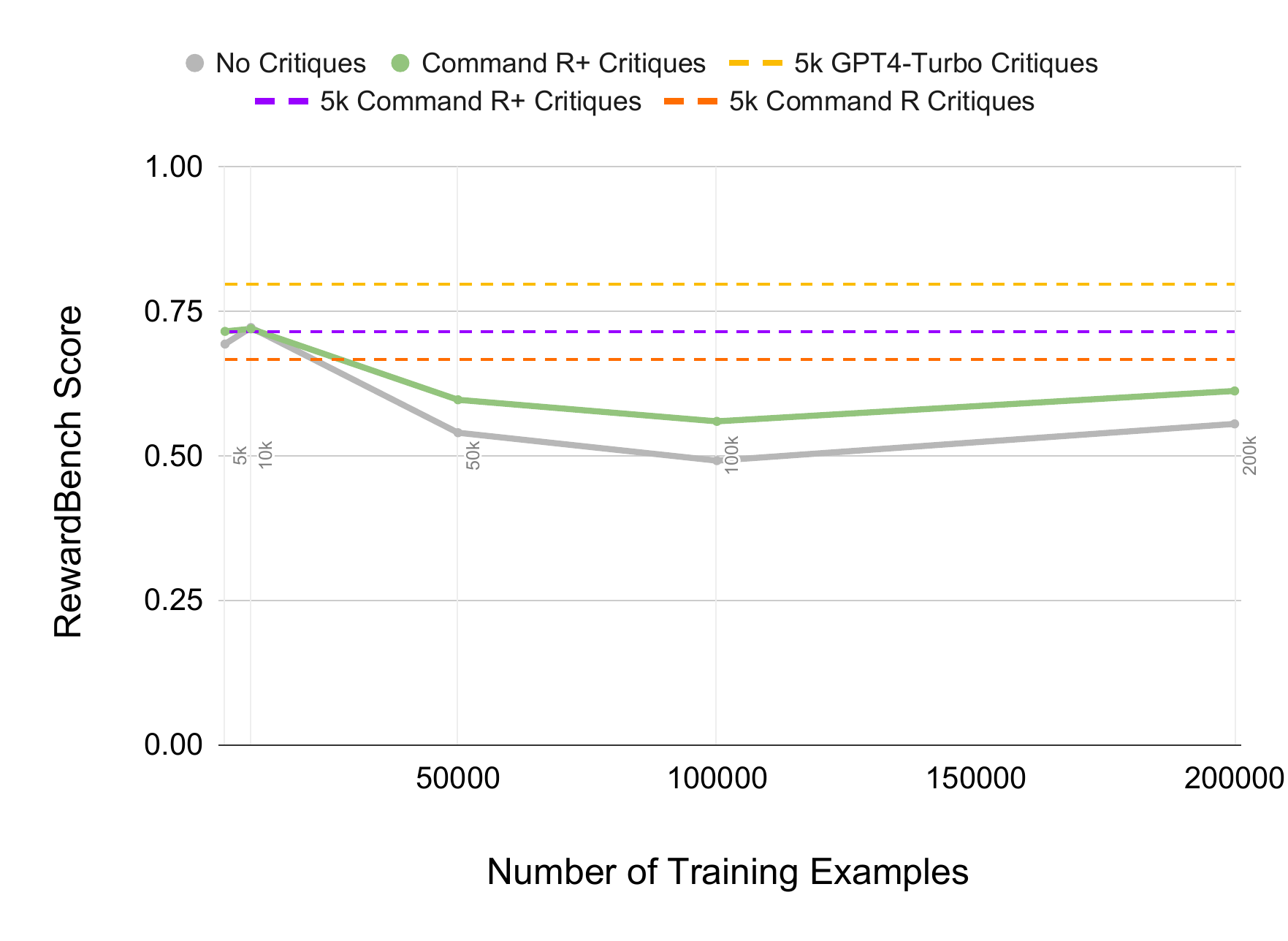}
        \caption{Chat Hard}
        \label{fig:rbsubsets_chathard}
    \end{subfigure}}
    \vskip\baselineskip
    \begin{subfigure}[b]{0.49\textwidth}
        \centering
        \includegraphics[width=\textwidth]{figures/chat.pdf}
        \caption{Chat}
        \label{fig:subplot3}
    \end{subfigure}
    \hfill
    \raisebox{0.0cm}{
    \begin{subfigure}[b]{0.49\textwidth}
        \centering
        \includegraphics[width=\textwidth]{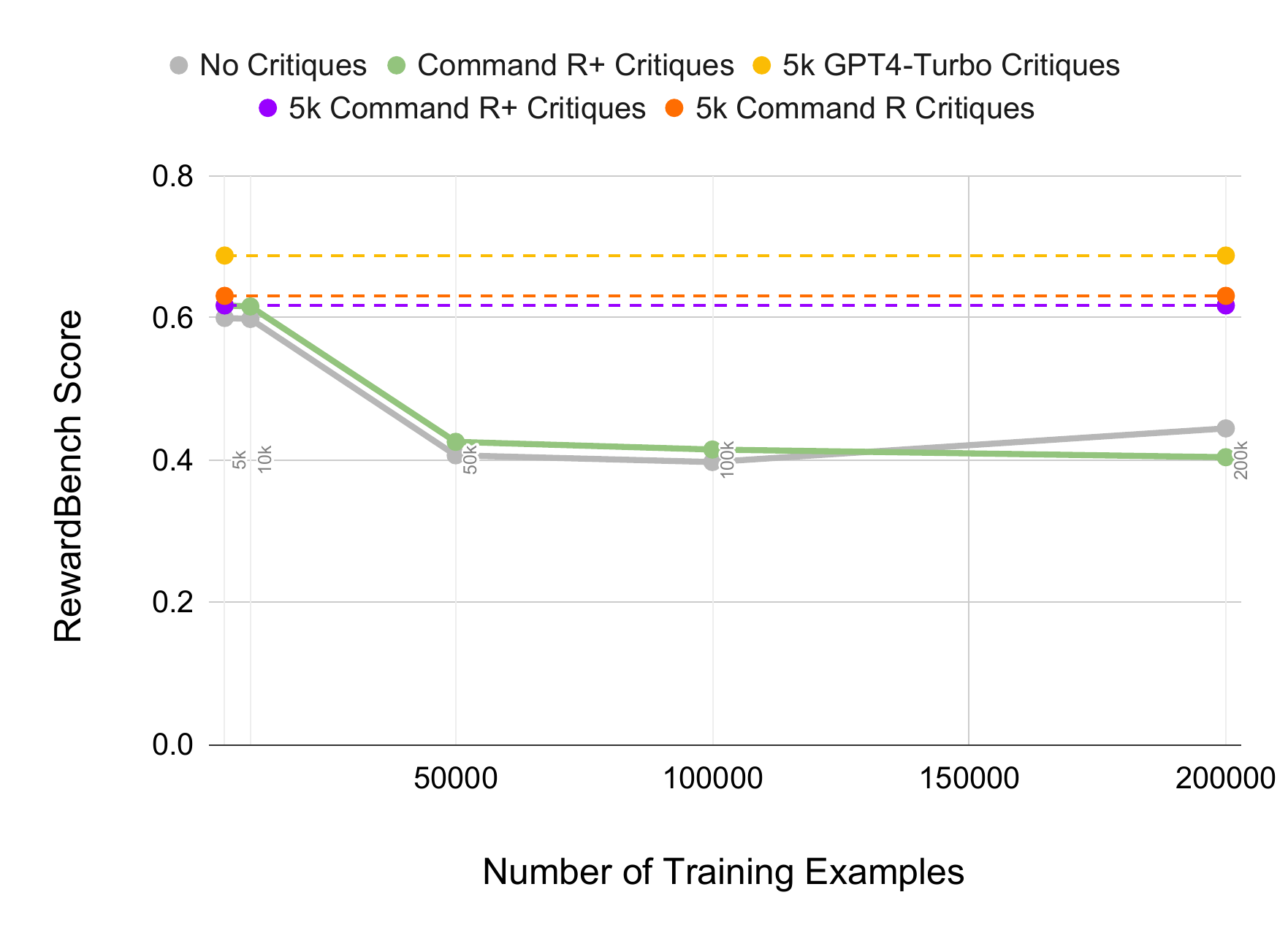}
        \caption{Safety}
        \label{fig:subplot4}
    \end{subfigure}}
    \caption{\textsc{RewardBench} Score v.s. training data size by task type.}
    \label{fig:task_scale}
\end{figure}

\section{Critique Generation Examples \label{app:critique_examples}}

Table~\ref{tab:case_ppo} has critiques from the different models for the same given (preferred) response.

\section{Scaling Performance By Task \label{app:performance_by_task}}
We illustrate a break down of \textsc{RewardBench} Score v.s. training examples by task type in Fig.~\ref{fig:task_scale}. We observe that RMs enriched with Command R+ Critiques consistently outperforms No-Critiques on \textit{Reasoning} task. \textit{Chat Hard} contains challenging prompt completions that are designed with superficial features to trick LLM-based evaluators. On this dataset, training with Command R+ Critiques leads to a less performance drop compared to No-Critiques over increasing number of training examples, suggesting that critiques help with robustness against adversarial examples. We observe that the classifier-based RMs overwhelmingly outperform the generative judge baseline. While the generative judge performs strongly on \textit{Chat}, comparably on \textit{Safety}, it performs poorly in \textit{Reasoning}, followed by \textit{Chat Hard}. Upon closer inspection on \textit{Reasoning}, we see that many of the mistakes for Gen-Command R+ (score of $~0.06$) are due to the generative judge assigning the same score to both preferred and worse, leading to a tie. We hypothesize that this is because without preference training, completion pairs from the more difficult \textit{Reasoning} subset appear similar to the pretrained checkpoint, which is not sensitive to their differences. However, when conditioning on critiques, Gen-Command R+ (\textit{+Critiques}) spreads out the reward scores on those ties, leading to a $0.3$ increase (score of $0.3685$).

\section{Critique Quality Analysis \label{app:critique_quality}}
We analyze the quality of the critiques generated by different models using metacritique scores proposed in~\citet{sun2024critique}, as shown in Table~\ref{tab:critique_quality}. ~\citet{sun2024critique} develop metrics to quantify the critique ratings using GPT4-generated critiques as reference. Specifically, \textit{precision} gauges the accuracy of the critique's content, ensuring each information unit contained in the critique is factual. \textit{Recall} measures the comprehensiveness of the critique, and the extent to which it fully covers the necessary breadth of information in the reference. $F1$ is the harmonic mean of \textit{precision} and \textit{recall}. For more detail please refer to \citet{sun2024critique}. We observe that the quality correlates to the size of the critique models. LLaMA2-7B-Chat shows relatively low metacritique scores (a Meta-$F1$ of 0.5481). Mixtral-8x7B-Instruct critiques are more comprehensive, whereas those generated by Command R families are more accurate. In the early stage of our experiments, we also trained on a small amount of human-annotated critiques and observed improved test accuracy slightly higher than GPT4-Turbo, validating that human critiques are helpful and effective in improving RM performance.      

\begin{table}[h!]
\centering
\small
\begin{tabular}{lrrrl}
\toprule
\textbf{Critiques} & {\textbf{Meta-Precision}} &  {\textbf{Meta-Recall}} & {\textbf{Meta-F1}} &  \\
\midrule
LLaMA2-7B-Chat & 0.6295 & 0.5504 & 0.5481 \\
Command R   & 0.7108 &	0.6005 &  0.6250  \\ 
Mixtral-8x7B-Instruct  & 0.7095	& 0.6553 & 0.6439 \\ 
LLaMA2-70B-Chat  & 0.6934  & 0.6184 & 0.6232 \\
Command R+   & 0.7667 &	0.6065 & 0.6422\\
\bottomrule
\end{tabular}
\caption{Metacritique scores of critiques generated by different models. }
\label{tab:critique_quality}
\end{table}

\begin{figure}[h!]
 \centering \includegraphics[width=0.90\linewidth]{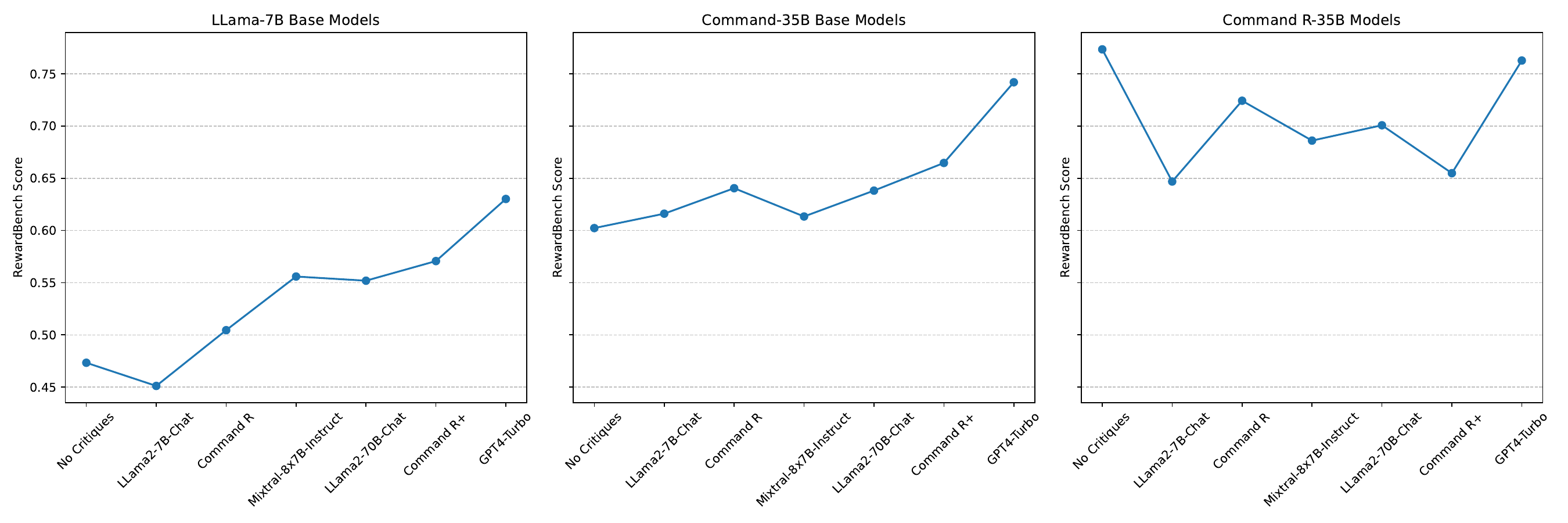}
 \vspace*{-5pt}
 \caption{\textsc{RewardBench} Score vs. critiques generated by models of increasing sizes. The $x$-axis is listed in increasing model parameter count. The effects of critiques on test accuracy correlates with critique model sizes, for base model checkpoints. \label{fig:rewardbench_vs_sizes}}
\end{figure}

\end{document}